# Natural Language Generation Using Link Grammar for General Conversational Intelligence

Vignav Ramesh[1,2] and Anton Kolonin[2,3,4]

[1] Saratoga High School, Saratoga CA 95070, USA
[2] SingularityNET Foundation, Amsterdam, Netherlands
[3] Aigents, Novosibirsk, Russian Federation
[4] Novosibirsk State University, Russian Federation
{rvignav, akolonin}@gmail.com

**Abstract.** Many current artificial general intelligence (AGI) and natural language processing (NLP) architectures do not possess general conversational intelligence—that is, they either do not deal with language or are unable to convey knowledge in a form similar to the human language without manual, labor-intensive methods such as template-based customization. In this paper, we propose a new technique to automatically generate grammatically valid sentences using the Link Grammar database. This natural language generation method far outperforms current state-of-the-art baselines and may serve as the final component in a proto-AGI question answering pipeline that understandably handles natural language material.

## 1 Introduction

### 1.1 Artificial General Intelligence and Natural Language Processing

Currently, the fields of AGI and NLP have little overlap, with few existing AGI architectures capable of comprehending natural language and nearly all NLP systems founded upon specialized, hardcoded rules and language-specific frameworks that are not generalizable to the various complex domains of human language [1]. Furthermore, while some NLP and AGI frameworks do incorporate some form of natural language understanding, they are still unable to convey knowledge in a form similar to the human language without manual, labor-intensive methods such as template-based customization.

### 1.2 Motivations

**Interpretable Language Processing.** Unlike explainable artificial intelligence (XAI), which refers to methods and techniques in the application of artificial intelligence (AI) such that the results generated by the AI model can be understood by humans, inter-pretable AI (IAI) requires both the AI model itself as well as its results to be understandable. It contrasts with the concept of "black box" algorithms in machine learning where even developers cannot explain how the AI arrived at a specific decision. Interpretable language processing (ILP) is an extension of the IAI concept to NLP; ILP is expected to allow for acquisition of natural language, comprehension of textual communications, and production of textual messages in a reasonable and transparent way [2]. Our natural language generation (NLG) architecture intends to serve as an ILP method that allows for the generation of grammatically valid sentences in an interpretable manner. Our work not only provides explainable results but also provides an interpretable model for sentence generation, since we rely on Link Grammar which is comprehensible in itself.

**Unsupervised Language Learning.** Current methods of grammar learning, such as deep neural networks (DNNs), require extensive supervised training on large corpora. However, humans are capable of acquiring explainable and reasonable rules of building sentences from words based on grammatical rules and conversational patterns and clearly understand the grammatical and semantic categories of words. This raises the idea of unsupervised language learning (ULL), which enables acquisition of language grammar from unlabeled text corpora programmatically in an unsupervised way. In a ULL system, the learned knowledge is stored in a human-readable and reasonable representation [2]. Examples of ULL systems include the OpenCog AI platform's cognitive pipeline that enables unsupervised grammar learning [3] and Glushchenko et al.'s grammar induction technology that unsupervisedly learns grammar from only a single input corpus [4].

The ULL pipeline consists of five main components: a text pre-cleaner (which preprocesses corpus files with configurable cleanup and normalization options), a sense pre-disambiguator (which performs word disambiguation and builds senses from tokens), a text parser (which parses sentences into word tokens or senses), a grammar learner (which learns word categories and rules from parses), and a tester that evaluates the quality of the inferred grammar. The ultimate output of the ULL pipeline is a model of the human language; one such model is the Link Grammar dictionary itself, which serves as the knowledge base behind our NLG architecture [5].

**Question Answering.** Question answering is a computer science discipline within the fields of information retrieval and natural language processing concerned with building systems that automatically answer questions posed by humans in a natural language. An explainable question answering pipeline would consist of two main components: natural language comprehension, and natural language generation. Natural language comprehension involves first parsing the question (otherwise known as the input query) based on a formal grammar such as Link Grammar, and then performing semantic interpretation (extracting the concept represented by the query and determining relationships between individual parts of the parsed query obtained in the previous step). The natural language generation component then involves semantic query execution (determining the answer to the input question based on the semantic relationships extracted in the previous step) and the use of a formal grammar to construct grammatically valid sentences from the words associated with the semantic relationships derived during query execution [2]. A diagram of the question answering pipeline is shown below.

# Question Answering Pipeline

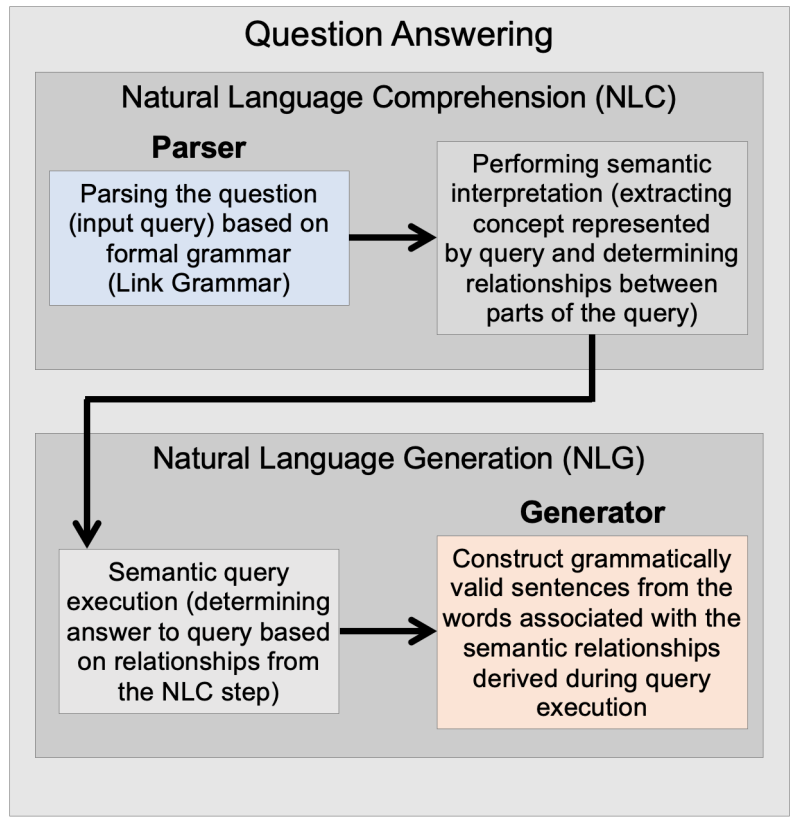

**Fig. 1.** Question answering workflow

Our current work is concerned with the final part of the question answering pipeline; with Link Grammar serving as the requisite formal grammar, the proposed NLG architecture automates sentence construction from the words derived in the penultimate part of the pipeline and thereby enables question answering with minimal to no human interference.

## 1.3   Link Grammar

Link Grammar is founded upon the idea that each word possesses a feature structure consisting of a set of typed connectors and disjuncts associating those connectors.

Rules, which correspond to lexical entries or grammatical categories, describe a list of words and a set of their defining disjuncts. Generation involves matching up connectors from one word with connectors from another, given the known set of disjuncts for both of them.

A connector represents either the left or right half of a grammatical link of a given type, and different types of connectors are denoted by letters or pairs of letters like S or SX. For instance, if a word A has the connector S+, this means that A can have an S link to its right. If a word B has the connector S-, this means that B can have an S link to its left. In this case, if A occurs to the left of B in a sentence, then the two words can be joined together with an S link [1]. The "Link Grammar dictionary," then, is a database that maps all common words of a given language to the connectors that define them.

Disjuncts are sets of connectors that constitute the legal use of a given word. For instance, if word A has the expression {C-} & (X+ or Y+), where {} denotes optionality, then A has the following four disjuncts: C- X+, X+, C- Y+, Y+.

Macros are single symbols that define large connector expressions. For instance, let the macro *<macro>* define the expression {C-} & (X+ or Y+) as above. Then, the expression *<macro>* or {W+ & Z-} is simply shorthand for ({C-} & (X+ or Y+)) or {W+ & Z-}. Macros are used throughout the Link Grammar dictionary to reduce redundancy of connector expressions and increase readability.

In Link Grammar notation, a sentence is a set of tokens (words, punctuation, and other syntactic structures) that may be linked by matching connectors between pairs of tokens. For instance, consider the sentence, "The cat caught a mouse." The Link Grammar parse structure for this sentence is:

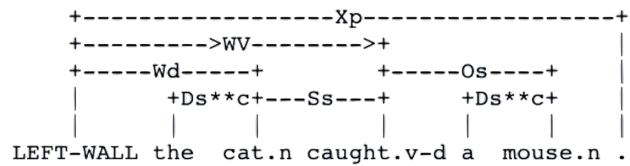

**Fig. 2**. Link Grammar parse of "The cat caught a mouse."

The rules of Link Grammar impose additional constraints beyond the matching of connectors, namely the planarity and connectivity metarules. Planarity means that links do not cross, while connectivity means that the links and words of a sentence must form a connected graph—all the words must be linked to the other words in the sentence via at least one path.

Overall, the structure of grammatical knowledge as stored in Link Grammar appears interpretable, human-readable, and maintainable, including the ability to add new rules and amend existing ones based on the intent of a human linguist maintaining the knowledge base or a programmatic language learning framework such as the ULL pipeline discussed in Section 1.2 [3, 4]. Our proposed architecture utilizes the English Link Grammar dictionaries to perform natural language generation.

**Link Grammar Database.** The Link Grammar database is comprised of dictionaries for each of more than 10 languages. In this paper, we focus on the English database, which includes approximately 1,430 distinct rules that each correspond to word clusters—groups of words with the same sets of grammar properties—and 86,863 word forms. Dictionaries may be represented as hierarchical trees of files, where a master ".dict" file maps all common words to their defining connector expressions and references supporting files that contain additional word clusters or their corresponding representation in a relational database [2].

**Why Link Grammar?** Besides Link Grammar, other grammar rule dictionaries and APIs exist, such as spaCy[1] and Universal Dependencies[2] (UD). Solutions such as spaCy are programming language-specific APIs or packages; for instance, spaCy restricts its grammar database to applications developed using Python. Moreover, unlike both spaCy and UD, which rely on or are structured as dependency grammars that require head-dependent relationships (links must be directional), Link Grammar does not mandate head-dependency. For example, consider the UD parse for the sentence "The dog was chased by the cat." below, which contains directional links:

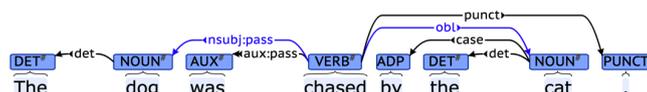

**Fig. 3**. UD parse of "The dog was chased by the cat."

As shown in Fig. 3., each link in the UD parse contains an arrow directed from the head component to the dependent component of the link. Link Grammar makes this directionality optional and can thereby be applied to a greater variety of sentential forms [6, 7].

More importantly, spaCy and similar libraries require grammar rules to be hardcoded into the client architectures for which they are utilized while Link Grammar does not, allowing continual updates and enhancement to the Link Grammar database (likely in the form of automated ULL input) without any client-side modifications. The human-readable and editable nature of Link Grammar allows our grammar induction algorithm to better serve as an INLP method for the purposes of the NLG task.

### 1.4    Natural Language Generation

Natural language generation refers to the process of mapping some representation of data or information (usually semantic and non-linguistic) to a linguistic (grammatically and morphologically valid) representation of that information. The task of generating even simple sentences demands a large body of knowledge (grammatical, syntactical, morphological, phonological, etc.). One possible way to build this knowledge base is to develop an integrated representation of semantic and grammatical knowledge using an extended Link Grammar schema—however, while this may be a product of our future work, our current NLG architecture is concerned only with generating sentences using the basic Link Grammar database in its current form.

---

[1]   `https://spacy.io`

[2]   `https://universaldependencies.org/introduction.html`

Our proposed architecture focuses on the surface realization component of NLG. Surface realization refers to a determination of how the underlying content of a text, in the form of a list of words, should be mapped into a sequence of grammatically correct sentences [8]. Often, the same content can be expressed in various sentential forms; an NLG system must first, determine which of these sentential forms is most appropriate, and second, ensure that the resulting sentence is syntactically and morphologically correct.

## 1.5    Prior Work

When considering small application domains with minimal variation, sentence generation is relatively simple, and outputs can be specified using templates (Reiter et al., 1995; McRoy et al., 2003). While templates allow for full control over the quality of the output and avoid the generation of ungrammatical structures, manually constructing them is a tedious and labor-intensive task. Most importantly, however, they do not scale well to applications involving heavy linguistic variation [9].

The KPML system (Bateman, 1997), based on Systemic-Functional Grammar (Halliday & Matthiessen, 2004), models surface realization as a traversal of a network in which the ideal route depends on both grammatical and semantic-pragmatic information. However, the high complexity and level of detail of these systems prevents them from being used as "plug-and-play" or "off the shelf" architectures (e.g., Kasper, 1989). A state-of-the-art example of these "plug-and-play" models is Dathathri et al.'s Plug and Play Language Model (PPLM), which combines a pretrained transformer-based language model (LM) with attribute classifiers that allow for controllable text generation. The flexibility of these attribute classifier combinations that can be plugged into the model (hence the name "plug-and-play") allow for various manners of guiding text generation and, in turn, a diverse range of NLG applications [10].

Lian et. al. proposed a natural language generation system via satisfaction of the constraints posed by inverse relations of hypergraph homomorphisms. To perform surface realization, the proposed OpenCog NLGen software uses the SegSim approach to take an Atom set in need of linguistic expression and match its subsets against a data-store of (sentence, link parse, RelEx relationship set, Atom set) tuples, produced via applying OpenCog's natural language comprehension tools to a corpus of sentences. Via this matching, it determines which syntactic structures have been previously used to produce relevant Atom subsets. It then pieces together the syntactic structures found to correspond to its subsets, thereby forming overall syntactic structures corresponding to one or more sentences. The sentence is solved for as a constraint satisfaction problem from the Atom set semantics [1]. While SegSim processes words unproblematically for relatively simple sentences, it becomes unreliable for sentences involving conjunctions or other complex syntactic forms.

Ratnaparkhi proposed three trainable systems for surface natural language generation. The first two systems, called NLG1 and NLG2, require a corpus marked only with domain-specific semantic attributes, while the last system, called NLG3, requires a corpus marked with both semantic attributes and syntactic dependency information. All systems attempt to produce a grammatical natural language phrase from a domain-specific semantic representation. NLG1 serves a baseline system and uses phrase frequencies to generate a whole phrase in one step, while NLG2 and NLG3 use maximum entropy probability models to individually generate each word in the phrase. The systems NLG2 and NLG3 learn to determine both the word choice and the word order of the phrase [11]. Not only are Ratnaparkhi's black box NLG systems restricted to certain domain-specific grammar representations, but, as supervised models, they require labeled training data, thus straying from the goal of building an NLG system that satisfies the concepts of ULL and ULP.

Wen et al. proposed a statistical language generator based on a semantically controlled Long Short-term Memory (LSTM) structure. The LSTM generator learns from unaligned data by jointly optimizing sentence planning and surface realization using a simple cross entropy training criterion, and language variation is achieved by sampling from output candidates [12]. However, like Ratnaparkhi's work, the LSTM structure inherently requires large amounts of labeled training data and is a black box algorithm, and hence is neither unsupervised nor interpretable.

Freitag and Roy proposed an NLG technique whereby, without any supervision and only based on unlabeled text, denoising autoencoders are used to construct a sentence from structured data interpreted as a "corrupt representation" of the desired output sentence. Freitag and Roy's model also extends to NLG for unstructured data in that their denoising autoencoder can generalize to unstructured training samples to which noise has previously been introduced [13].

## 2    Methodology

Our NLG architecture consists of two main components: the Loader, and the Generator. The Loader is simply a utility program used by the Generator to store Link Grammar in memory; it is not specific to our NLG architecture but rather a tool for loading Link Grammar that can be used by any application [2]. Fig. 4. displays the presented NLG architecture, including the workflow of Loader and Generator.

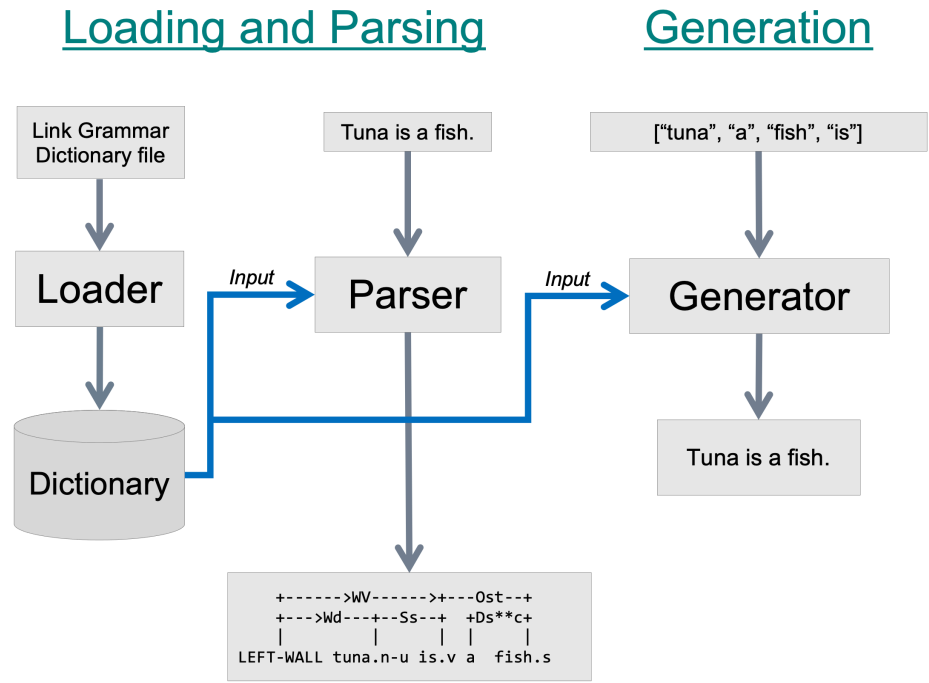

**Fig. 4.** Overall NLG architecture involving a sentence parsing algorithm (the "Parser") and question answering framework built upon the Generator, all relying on the same Link Grammar dictionary Loader infrastructure.

## 2.1    Loader

The Loader is responsible for loading and storing the Link Grammar dictionary into memory for future usage [2]. Namely, the Dictionary class stores a list of Word objects, where each word corresponds to a distinct Rule object; a Rule object may associate multiple Word objects and contains a list of Disjunct objects that each contain a list of connectors constituting the legal use of any of the words that correspond with the given rule. Currently, the Loader architecture only supports English and does not handle complex morphology structures such as those needed to support languages that require heavy morphology usage (one such language being Russian).

   The core of the Loader is the makeDict() function, which converts an array of lines (obtained from the Link Grammar database) into an array of Dictionary objects containing the rules for all common words and phrases that Link Grammar provides support for. The basic framework of makeDict() is as follows:

---

**Algorithm 1:** MAKEDICT

**Input** : An array *lines* of all lines in the Link Grammar Database
**Output:** An array [*dict*, *hyphenated*] of Dictionary objects, one for common words and one for common phrases with words separated by underscores

Initialize *dict* and *hyphenated*
Initialize *macros*, which maps single links to the large connector expressions they define

**define** ASSIGN(*w*, *r*):
**if** *w is a hyphenated phrase* **then**
  | Add (*w*, *r*) to *hyphenated*
**else**
  | Add (*w*, *r*) to *dict*
**end**
**end**

**for** *line in lines* **do**
  **if** *line starts with a macro* **then**
    Split the single link, *macro*, from its definition, *rule*
    Add (*macro*, *rule*) to *macros*
  **else**
    **if** *line contains a filename f* **then**
      Parse *f* to obtain the list of words it contains
      Replace all instances of macros in the rule *rule* specified in the following lines
       of the Link Grammar database with their expanded definitions as contained in
       *macros*
      Store *rule* in a Rule object *r*
      **for** *word w in f* **do**
        | ASSIGN(*w*, *r*)
      **end**
    **else**
      Split the word, *w*, from its definition, *rule*
      Process *rule* and store it in a Rule object *r*
      Replace all instances of macros in *rule* with their expanded definitions as
       contained in *macros*
      ASSIGN(*w*, *r*)
    **end**
  **end**
**end**
**return** [*dict*, *hyphenated*]

---

**Algorithm 1**. makeDict() function

As shown in the above algorithm, makeDict() parses the lines of the Link Grammar database to assign rules (within which all macros are expanded) to the words that they define. After splitting the contents of the Link Grammar database file, the Loader calls makeDict() to obtain the Dictionaries that store the Link Grammar rules for use in the Generator.

## 2.2 Generator

After obtaining the loaded Dictionary objects from the Loader, the Generator determines what sentences can be formed from a given list of words via valid Link Grammar rules. First, given a list of words, the Generator determines a subset of all orderings of those words that satisfies initial checks of the planarity and connectivity metarules (for example, one partial connectivity check that the Generator implements involves ensuring that the first and last words are capable of forming links to the right and left, respectively). Then, for each ordering in the subset, the Generator determines if that ordering is valid; specifically, it ensures that every pair of consecutive words (with certain exceptions as discussed below) can be connected via links part of the Dictionary objects. To do so, the Generator uses the connects() function, which returns a boolean value indicating whether its two parameters, the tokens *left* and *right*, can be linked together:

---
**Algorithm 2:** CONNECTS

  **Input** : A pair of strings $left$ and $right$, representing the two words to potentially be connected

  **Output:** An boolean value indicating whether $left$ and $right$ can be connected via valid Link Grammar rules

  Obtain $leftList$, the list of rules corresponding with $left$ (i.e. the rule when $left$ is a verb, the rule when $left$ is a gerund, etc.), from the global Dictionary variables $dict$ and $hyphenated$
  Obtain $rightList$ in a similar manner

  **for** $leftRule$ in $leftList$ **do**
    **for** $rightRule$ in $rightList$ **do**
      Split $leftRule$ and $rightRule$ into lists of Disjuncts $ld$ and $rd$
      **for** $l$ in $ld$ **do**
        **for** $r$ in $rd$ **do**
          Replace all instances of '−' in $l$ with '+' and vice versa
          **if** $l = r$ **then**
            **return true**
          **else**
            **continue**
          **end**
        **end**
      **end**
    **end**
  **end**
  **return false**
---

**Algorithm 2**. connects() function

As shown in the above algorithm, connects() first obtains the lists of rules leftList and rightList corresponding with left and right, and then checks if any Disjunct in any Rule in leftList matches with any Disjunct in any Rule in rightList.

However, connects() is not always applicable. For instance, when the determiner "a" is present in the phrase "is a human," the links are not "is" → "a" and "a" → "human" but rather "is" → "human" and "a" → "human" as shown in the Link Grammar parse below:

```
    +---Osm--+
    |  +Ds**c+
    |  |    |
  is.v  a  human.n
```

**Fig. 5**. Link Grammar parse of "is a human"

Hence, functions similar to connects() are implemented that deal with specific cases involving links not between consecutive words but rather from the far left or far right words to each of the other words in a set of three or more consecutive words.

A sample surface realization query, along with its result as generated by our NLG architecture, is as follows:

    GENERATE(["mom", "dad", "company", "wants", "join", "the", "to"]) =
["mom wants dad to join the company", "dad wants mom to join the company"]

As seen here, the proposed architecture generates grammatically valid sentences, but succumbs to the grammatical ambiguity problem described in the following section.

# 3    Results

Our algorithm was tested on two distinct corpora in a total of three different tests. We found that the accuracy of our results was affected primarily by the issue of grammatical ambiguity, which refers to situations in which the same word may have different roles in a sentence (i.e. a noun can be either a subject or an object) or may represent different parts of speech (such as the word "saw" in its verb and noun forms). Subject-object ambiguity, a specific case of grammatical ambiguity, refers to the potential interchange-ability of the subject and object in a sentence. For instance, consider the sentence "The cat caught a mouse" as referenced in 1.3. Both "The cat caught a mouse" and "The mouse caught a cat" are grammatically valid sentences that can be constructed from the five words "the," "cat,", "caught," "a," and "mouse;" the presence of these extra sentences that are grammatically valid yet contextually wrong skewed the accuracy of our algorithm.

Grammatical disambiguation—a solution to the grammatical ambiguity problem—refers to the process of determining which sentential form is most appropriate to represent some non-linguistic content; for instance, the sentence "The cat caught a mouse" is likely more contextually valid than "The mouse caught a cat." Semantic (word sense) disambiguation refers to the process of determining which "sense" or meaning of a word is activated by the use of the word in a particular context; for instance, the word "saw", when used in the sentence "Mom saw a bird," will have a different sense—and thus different Link Grammar rules—than when it is used in the sentence "The lumber-jack is holding a saw." Implementing grammatical and semantic disambiguation will be a product of our future work.

Our algorithm was primarily tested on 92 sentences with words all part of Singular-ityNET's "small world" POC-English corpus.[3] For this purpose, we have used a corresponding "small world" Link Grammar dictionary (automatically inferred from high quality Link Grammar parses created by SingularityNET's ULL pipeline) containing 42 total words and 5 total word clusters.[4] We use the following metrics to evaluate our proposed architecture:

***Architecture-Specific Metrics***:
- Single correct generated sentence: the number of times the algorithm generates a single candidate sentence that exactly matches the reference sentence
- Multiple sentences with one correct: the number of times the algorithm generates multiple candidate sentences, one of which exactly matches the reference sentence
- Multiple sentences with none correct: the number of times the algorithm generates multiple candidate sentences, none of which exactly matches the reference sentence
- No generated sentences: the number of times the algorithm fails to generate a candidate sentence

---

[3]    `http://langlearn.singularitynet.io/data/poc-english/poc_english.txt`

[4]    `http://langlearn.singularitynet.io/test/nlp/poc-english_5C_2018-06-06_0004.4.0.dict.txt`

- Too many results: the number of times the algorithm generates over 25 candidate sentences
- Accuracy: the proportion of results within the categories "Single correct generated sentence" and "Multiple sentences with one correct"

***Canonical NLG Metrics***:

- Average BLEU (Bigram) [14]: BLEU (Bilingual Evaluation Understudy) is calculated by counting the number of matching $n$-grams in the candidate and reference sentences. It is represented as a float between 0 and 1 inclusive, where values closer to 1 denote more similar texts. Given the short sentence length of SingularityNET's "small world" corpus, we use the bigram variant of BLEU, meaning that $n = 2$ for the above explanation.
- Average Word2Vec Cosine Similarity [15]: Word2Vec Cosine Similarity is calculated by first encoding each sentence as a vector, which is accomplished using the RoBERTa sentence transformer [16], and then calculating the cosine of the angle, or angular distance, between the vectors. Like BLEU, Word2Vec Cosine Similarity is represented as a float between 0 and 1 inclusive, where values closer to 1 denote more similar texts.
- Average WER [17]: WER (Word Error Rate) is a measure of the Levenshtein distance, or edit distance, for two sentences; in other words, it is a function the minimum number of edits (insertions, deletions, or substitutions) required to change the candidate sentence into the reference sentence. WER is calculated as $(S + D + I)/N$, where S is the number of necessary substitutions, D is the number of deletions, I is the number of insertions, and N is the total number of words. It takes on a minimum value of 0 (which means the sentences are identical) and has no maximum value (an arbitrary number of words can be inserted into the sentence); values closer to 0 indicate greater similarity.
- Average TER [18]: TER (Translation Edit Rate) refers to the number of edits required to make a candidate sentence exactly match its reference sentence in fluency and semantics. It is calculated as $E/R$, where E is the minimum number of edits and R is the average length of the reference text. Like WER, its minimum is 0 but it has no maximum value, and TER scores closer to 0 denote more similar texts.

By "average," we mean that the given metric is calculated for each pair of candidate and reference sentences and then averaged over the entire corpus.

**Table 1.** Results when tested on 92 sentences with words from SingularityNET's "small world" corpus using ULL-generated grammar.

| Metric | Result |
|---|---|
| ***Architecture-Specific Metrics*** | |
| Single correct generated sentence | 62/92 |
| Multiple sentences with one correct | 30/92 |
| Multiple sentences with none correct | 0/92 |
| No generated sentences | 0/92 |
| Too many results | 0/92 |
| Accuracy | 1.000 |
| | |
| ***Canonical NLG Metrics*** | |
| Average BLEU (Bigram) | 1.000 |
| Average Word2Vec Cosine Similarity | 0.988 |
| Average WER | 0.246 |
| Average TER | 0.082 |

When tested on the same 92 sentences while using the complete Link Grammar dictionary for English,[5] the algorithm achieved the following results:

**Table 2.** Results when tested on 92 sentences with words from SingularityNET's "small world" corpus using complete Link Grammar.

| Metric | Result |
|---|---|
| ***Architecture-Specific Metrics*** | |
| Single correct generated sentence | 8/92 |
| Multiple sentences with one correct | 57/92 |
| Multiple sentences with none correct | 0/92 |
| No generated sentences | 0/92 |
| Too many results | 27/92 |
| Accuracy | 0.707 |
| | |
| ***Canonical NLG Metrics*** | |
| Average BLEU (Bigram) | 0.999 |
| Average Word2Vec Cosine Similarity | 0.900 |
| Average WER | 3.713 |
| Average TER | 0.395 |

The decreased value of the "Single correct generated sentence" metric, increased value of the "Multiple sentences with one correct" metric, slightly decreased BLEU, decreased Word2Vec Cosine Similarity, increased WER, and increased TER in the second test as compared to first test are all direct results of the increased grammatical and semantic ambiguity caused by using Link Grammar instead of SingularityNET's "small world" grammar. Since the "small world" grammar was created from the "small world" corpus itself, each of the words in the corpus contains no other grammatical or semantic contexts (and thus no other sets of grammar rules) besides those required to form the 92 sentences that the algorithm was tested on. However, since the Link Grammar database is much larger and contains many more "senses" for each word, there is a greater number of valid sentences that can be constructed from those words, thereby causing a majority of reference sentences previously associated with a single candidate sentence in the first test to instead correspond with several candidate sentences in the second test.

---

[5] `https://github.com/opencog/link-grammar/tree/master/data/en`

Our NLG architecture was also tested on 54 sentences part of Charles Keller's production of Lucy Maud Montgomery's "Anne's House of Dreams" as found in the Gutenberg Children corpus,[6] and performed as follows:

**Table 3.** Results when tested "Anne's House of Dreams" using complete Link Grammar.

| Metric | Result |
|---|---|
| ***Architecture-Specific Metrics*** | |
| Single correct generated sentence | 1/54 |
| Multiple sentences with one correct | 53/54 |
| Multiple sentences with none correct | 0/54 |
| No generated sentences | 0/54 |
| Accuracy | 1.000 |
| | |
| ***Canonical NLG Metrics*** | |
| Average BLEU (Bigram) | 0.652 |
| Average Word2Vec Cosine Similarity | 0.746 |
| Average WER | 5.976 |
| Average TER | 1.738 |

Here, the decreased BLEU and Word2Vec Cosine Similarity as well as increased WER and TER are due to the increase in results satisfying the "Multiple sentences with one correct" metric. Not only is the proportion of results in this category greater, but the number of candidate sentences generated for each reference sentence is also much greater due to longer average sentence length and the increased number of "senses" per word used in the Gutenberg Children corpus, resulting in lower average similarity scores and greater average error and edit rates.

Our NLG architecture outperforms prior work. As a baseline, we implemented a (slightly modified) version of the state-of-the-art Transformer model proposed in [19] for the task of sentence reconstruction from randomly shuffled sets of tokens, which exactly matches the surface realization task that our proposed NLG algorithm accomplishes (note that the algorithms described in Section 1.5 did not contain publicly available code or model schematics and thus could not be used as baselines). The implemented Transformer, which is built upon spaCy's English grammar, differs from [19] in that it uses a learned positional encoding rather than a static one, label smoothing is not utilized, and a standard Adam optimizer with a static learning rate is used instead of one with warm-up and cool-down steps (the reason for making these changes is that the modified Transformer closely matches the majority of current Transformer variants, such as Bidirectional Encoder Representations from Transformers, or BERT) [20]. This baseline implementation achieved the following results on the "small world" POC-English corpus and the "Anne's House of Dreams" excerpt of the Gutenberg Children corpus:

---

**Table 4.** Baseline results on the POC-English and Gutenberg Children corpora.

| Corpus / Metric | Result |
| --- | --- |
| *POC-English Corpus* | |
| Average BLEU (Bigram) | 0.747 |
| Average Word2Vec Cosine Similarity | 0.722 |
| Average WER | 3.114 |
| Average TER | 0.505 |
| | |
| *Gutenberg Children Corpus* | |
| Average BLEU (Bigram) | 0.325 |
| Average Word2Vec Cosine Similarity | 0.401 |
| Average WER | 11.622 |
| Average TER | 1.988 |

Our NLG architecture, when using complete Link Grammar (and when using SingularityNET's ULL-generated grammar for the POC-English corpus), far outperformed the baseline model in terms of BLEU, Word2Vec Cosine Similarity, and TER; it achieves a competitive WER score on the POC-English corpus when using complete Link Grammar and a far superior WER score on the POC-English corpus when using the ULL-generated grammar and on the Gutenberg Children corpus with complete Link Grammar. Overall, we find that our proposed architecture outperforms the Transformer—a ubiquitous state-of-the-art NLG model—and achieves stronger results across the board.

## 4 Conclusion

Our NLG architecture, when paired with the requisite task-specific NLP algorithms, may be used in a variety of scenarios. It can primarily be applied to the question answering problem as discussed in 1.2.; one such application is the Aigents Social Media Intelligence Platform [21]. Currently, the Aigents framework relies on artificially designed controlled language resembling oversimplified "pidgin" English; our proposed NLG algorithm can be integrated into the Aigents cognitive architecture to provide Aigents with full conversational intelligence [22]. In this scenario, natural language text would be produced in a quality higher than that provided by Aigents' current chatbots while also being explainable.

Another prominent subdomain of question answering in which our NLG architecture can be utilized is that of virtual assistant AI technologies such as Alexa and Google Home. The proposed NLG system, if integrated into these technologies in conjunction with the other parts of the question answering pipeline, would serve as an understandable and unsupervised rather than black box NLP framework.

Our further work will be dedicated to: 1) implementing grammatical and semantic disambiguation based on the context of a sentence or body of text; and 2) extending the algorithm's generation capabilities to languages other than English (including those that require heavy morphology usage).

# 5    Code Availability

Our NLG architecture is open-source and available under the MIT License (a permissive, limited-restriction license) on GitHub at `https://github.com/aigents/ai-gents-java-nlp`.